\documentclass[lettersize,journal]{IEEEtran}
\usepackage{amsmath,amsfonts}
\usepackage{array}
\usepackage[caption=false,font=normalsize,labelfont=sf,textfont=sf]{subfig}
\usepackage{textcomp}
\usepackage{stfloats}
\usepackage{url}
\usepackage{verbatim}
\usepackage{graphicx}
\usepackage{amsmath} % for \boldsymbol macro
\usepackage{bm}      % for \bm macro
\usepackage{multirow}

\usepackage{algorithm}
\usepackage{algpseudocode}
\usepackage{amsmath}
\def\BibTeX{{\rm B\kern-.05em{\sc i\kern-.025em b}\kern-.08em
    T\kern-.1667em\lower.7ex\hbox{E}\kern-.125emX}}
\usepackage{balance}
\begin{document}

\title{\Large Generative AI Enables Structural Brain Network Construction \\
from fMRI via Symmetric Diffusion Learning}

\author{Qiankun Zuo, Bangjun Lei, Wanyu Qiu, Changhong Jing, Jin Hong, Shuqiang Wang
\thanks{Corresponding authors: Jin Hong(Email:hongjin@ncu.edu.cn), Shuqiang Wang(Email:sq.wang@siat.ac.cn)}
\thanks{Qiankun Zuo, Bangjun Lei and Wanyu Qiu are with the Hubei Key Laboratory of Digital Finance Innovation, the School of Information Engineering, and also the Hubei Internet Finance Information Engineering Technology Research Center, Hubei University of Economics, Wuhan 430205, Hubei China.}
\thanks{Changhong Jing is with the Department of Computing, Hong Kong Polytechnic University, Hongkong 999077, China.}
\thanks{Jin Hong is with the School of Information Engineering, Nanchang University, Nanchang 330031, Jiangxi, China.}
\thanks{Shuqiang Wang is with the Shenzhen Institutes of Advanced Technology, Chinese Academy of Sciences, Shenzhen, 518055,China. }
}

%\markboth{Journal of \LaTeX\ Class Files,~Vol.~18, No.~9, September~2020}%
%{How to Use the IEEEtran \LaTeX \ Templates}

\maketitle

\begin{abstract}
Modeling the complex structural-functional relationship between brain regions can help uncover the underlying pathological mechanisms during neurodegenerative disease development and progression. Mapping from functional connectivity (FC) to structural connectivity (SC) can facilitate multimodal brain network fusion and discover potential biomarkers for clinical implications. However, it is challenging to directly bridge the reliable non-linear mapping relations between SC and functional magnetic resonance imaging (fMRI). In this paper, a novel symmetric diffusive generative adversarial network-based fMRI-to-SC (DiffGAN-F2S) model is proposed to predict SC from brain fMRI in a unified framework. To be specific, the proposed DiffGAN-F2S leverages denoising diffusion probabilistic models (DDPMs) and adversarial learning to efficiently generate symmetric and high-fidelity SC through a few steps from fMRI. By designing the dual-channel multi-head spatial attention (DMSA) and graph convolutional modules, the symmetric graph generator first captures global relations among direct and indirect connected brain regions, then models the local brain region interactions. It can uncover the complex mapping relations between fMRI and symmetric structural connectivity. Furthermore, the spatially connected consistency loss is devised to constrain the generator to preserve global-local topological information for accurate symmetric SC prediction. Testing on the public Alzheimer's Disease Neuroimaging Initiative (ADNI) dataset, the proposed model can effectively generate empirical SC-preserved connectivity from four-dimensional imaging data and shows superior performance in SC prediction compared with other related models. Furthermore, the proposed model can identify the vast majority of important brain regions and connections derived from the empirical method, providing an alternative way to fuse multimodal brain networks and analyze clinical brain disease.
\end{abstract}


\section{Introduction}

\label{s1}
\IEEEPARstart{T}{he} human brain is the most complex and sophisticated biological system with its intricate and interconnected network of neurons, which describes the cognitive and behavioral dysfunctions during daily activities\cite{zhan2014deficient,deluca2020treatment}. These dysfunctions can be expressed by the brain network, which consists of billions of interconnected neurons. Disruption of brain networks can lead to the emergence of various brain diseases and disorders\cite{kapogiannis2011disrupted}, such as Alzheimer's disease (AD) and Parkinson's disease (PD). Each disorder exhibits unique patterns of disruption within the brain networks, resulting in diverse and often debilitating symptoms. Understanding how these diseases perturb the intricate communication pathways and functional dynamics of the brain network is crucial to developing targeted therapies and diagnostic tools that can revolutionize patient care.

The brain network is commonly divided into structural connectivity (SC) and functional connectivity (FC)\cite{park2013structural,chen2025connectomediffuser}. The SC, derived from diffusion tensor imaging (DTI), represents the physical pathways that link different brain regions through bundles of axonal fibers. By tracing these pathways, researchers gain valuable knowledge about the brain's anatomical organization, elucidating the framework within which information flows across distant brain regions\cite{fox2022association}. Besides, the FC derived from functional magnetic resonance imaging (fMRI) unveils the functional alliances and neural circuits that underlie various cognitive functions\cite{diez2018dynamic,fiorenzato2019dynamic}, such as memory, attention, and emotion. Either SC or FC can be explored to detect disease-related abnormal connections for clinical diagnosis and treatment\cite{chetelat2013relationships}, which has unique advantages in brain disease diagnosis compared to the method based on Euclidean features\cite{wang2020ensemble,chen2024ig}. Due to the power of multimodal fusion, fusing SC and FC enhances structural-functional connectivity complementarity and greatly improves the capabilities of disease analysis\cite{yu2020individualized,sui2021structural,gong2023generative}. For example, Lei \emph{et al.}\cite{lei2021auto} utilized an auto-weighted centralized multi-task method to combine FC and SC for selecting disease-related informative features and diagnosis improvement. Also, the work in \cite{yu2020multi} concatenated the SC and FC for each subject and applied a multi-scale graph convolutional network (GCN) for mild cognitive impairment detection by introducing non-imaging data. To exploit topological information among brain regions, Huang \emph{et al.}\cite{huang2019integrating} treated SC and FC as graphs and edges and built graph-based networks to extract multi-scale features by one- and two-step graph convolutional diffusions. Good performance has been achieved on the epilepsy diagnosis. In addition, Dsouza \emph{et al.}\cite{dsouza2021m} modified the GCN-based networks and designed row-column filters to fuse SC and FC for phenotypic characterization. Because of the high cost and time-consuming problems, both SC and FC are not easily accessible. Cross-modal mapping between them is an alternative way to solve this problem and facilitate the SC-FC fusion for disease analysis. Furthermore, constructing SC from brain fMRI can help explore the many-to-one relations from functional modality to structural modality.

\begin{figure}[htbp]
	\centering
	\includegraphics[width=\columnwidth]{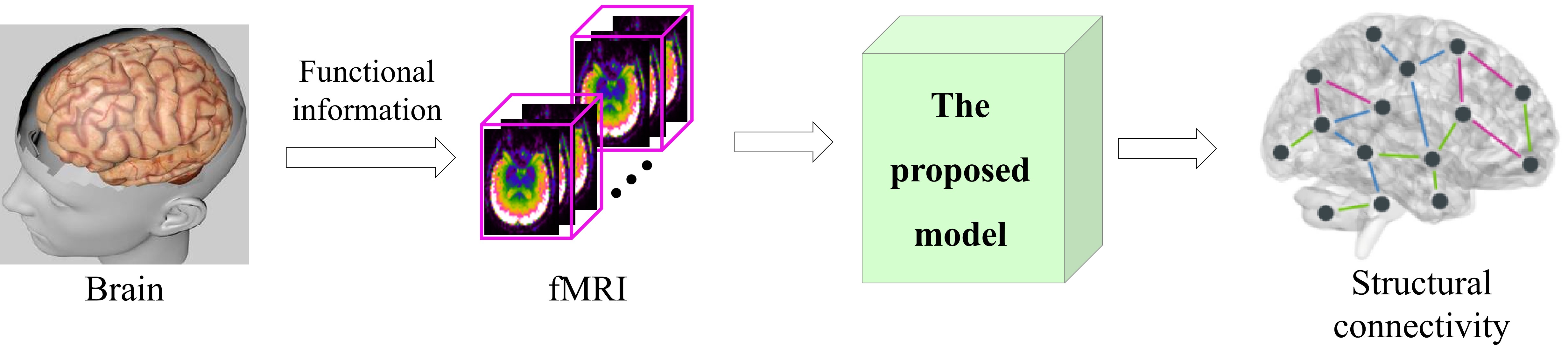}
	\caption{The flow chart of predicting structural connectivity from brain fMRI using the proposed method. \label{fig1}}
\end{figure}

The close relationship between structural connectivity and functional connectivity has been consistently validated and replicated across numerous studies. Previous works concentrated on constructing linear relationships for characterizing consistent patterns between SC and FC, such as correlation coefficients, degree properties, centrality measurements, and so on. Recently, deep learning has brought breakthroughs for modeling the relationship between SC and FC using three major approaches.
The first approach is to predict FC from SC\cite{honey2009predicting}. Previous studies have validated that brain structural connectivity plays a significant role in shaping brain functional patterns. Inspired by this literature, Zhang \emph{et al.}\cite{zhang2020deep} modeled non-linear higher-order mappings from SC to FC and learned informative brain saliency regions for PD analysis. Sarwar \emph{et al.}\cite{sarwar2021structure} investigated the structure-function coupling by predicting individual FC and proved a considerably stronger coupling than previously discovered.
Predicting static SC from dynamic FC is more effective and optimal, which is the second approach and has attracted increasing attention. For example, Wang \emph{et al.}\cite{wang2020understanding} removed the indirect and modular connections from FC and applied a deconvolution algorithm to predict SC with a highly consistent edge distribution. Considering the graph properties, Zhang \emph{et al.}\cite{zhang2022predicting} designed a multi-GCN-based GAN model to model complex relationships from FC to SC in topological space. Due to the fact that the presence of negative connections in FC holds valuable information for neural signal dynamics and physical structure, Tang \emph{et al.}\cite{tang2023signed} devised a signed graph encoder to learn cross-modal connectivity projection from FC to SC, which facilitates the detection of phenotypic and disease-related biomarkers and provides valuable support for biological interpretation. The third approach is to model a bidirectional mapping relationship between SC and FC\cite{bi2024cross,zuo2025bidirectional,tan2025connectome}, they constructed a unified framework and make the cross-modal brain network generation more efficient. However, these traditional approaches are based on two stages in predicting SC from raw brain fMRI, where the first step is to compute the FC from fMRI by the software, and the second step is to build deep learning models to predict SC from FC. It is inefficient and cannot be widely applied in clinical analysis. Besides, they heavily rely on manual parameter setting, which may cause large connectivity calculation errors and lower the performance of functional-to-structural connectivity prediction.

Generative Adversarial Networks (GANs)\cite{goodfellow2020generative} offer a compelling advantage in medical image computing due to their ability to effectively learn distribution-consistent features \cite{yu2022morphological,you2022fine,wang2022brain}. The cross-modal prediction computing allows researchers and healthcare professionals to leverage existing datasets from various modalities, enhancing data utilization and reducing the need for laborious and expensive data collection for each modality separately\cite{cai2019towards,hu2021bidirectional}. To predict SC from brain fMRI in one stage, the GAN-based model remains suitable and effective. The optimized generator maps the fMRI into structural connectivity in an end-to-end manner when the discriminator fails to distinguish the connectivity distribution. However, the challenge of instability and mode collapse during adversarial training still needs to be fully addressed.
Meanwhile, denoising diffusion probabilistic models (DDPM)\cite{ho2020denoising,nichol2021improved} have gained considerable attention in machine learning and demonstrated promising performance in image generation\cite{zong2024new}. DDPM utilizes a diffusion process to model the data distribution, gradually transforming Gaussian noise into a clean sample through a lot of successive steps\cite{zuo2025spatiotemporal,zuo2025brain}. Recent works\cite{wang2025generative,yao2025catd} of cross-modal imaging data generation have designed the DDPM-related model to achieve good results. The main advantage is the high quality and diversity of generation performance, which can address the problems of mode collapse and training instability in GAN-based models. But thousands of denoising steps result in low efficiency in generating samples. Therefore, combining GAN and DDPM is an alternative way to efficiently generate samples with high quality and diversity.

Based on the aforementioned insights, we propose DiffGAN-F2S, a symmetric and efficient denoising diffusion generative adversarial network transmitting fMRI to SC in an end-to-end manner. The main idea of this work is shown in Fig.~\ref{fig1}. The fMRI is integrated as a condition in the denoising process of DDPM for structural connectivity prediction, where each denoising step is accompanied by a generator and a discriminator.
Firstly, the prior anatomical knowledge is used to transform each 3D volume into multiple ROIs (Regions of Interest) and calculate the preliminary ROI-based time series. Secondly, the empirical SC is transformed into a Gaussian matrix with thousands of difusion steps.
Thirdly, the preliminary ROI-based time series and the noisy Gaussian matrix are sent into the symmetric graph generator for noise removal; the dual-channel multi-head spatial attention (DMSA) and GCN-based modules in the generator model global-local interactions between brain regions to capture topological connectivity patterns; meanwhile, the connectivity discriminator is used to constrain the distribution of denoised SC to follow the distribution of empirical SC.
After a few denoising steps, the proposed DiffGAN-F2S finally enables the precise and effective mapping of fMRI to structural connectivity. To summarize, the primary contributions are as follows:
\begin{itemize}
	\item The proposed model is the first work to translate brain fMRI into structural connectivity in the field of cross-modal image-to-graph prediction. By leveraging the unique strengths of diffusion models and generative adversarial networks, it achieves high-fidelity and diverse structural connectivity prediction with efficient generation through a few denoising steps.
	\item The symmetric graph generator is designed to denoise the fMRI guided by the symmetrical connectivity. Through the dual-channel multi-head spatial attention (DMSA) and GCN-based modules, the generator first captures global relations among direct and indirect connected brain regions, then models the local brain region interactions, which uncovers the complex mapping between fMRI and structural connectivity.
	\item The spatially connected consistency loss is devised to preserve both global and local topological characteristics, guiding the denoising process to accurately predict intrinsic structural patterns.
\end{itemize}

The structure of this work is outlined as follows: Section ~\ref{s3} presents the main architecture of the proposed DiffGAN-F2S model. In Section ~\ref{s4}, we apply the proposed model to public datasets and conduct result analysis. Section ~\ref{s5} and ~\ref{s6} demonstrates the reliability of our findings and summarizes the key observations in this study.

\begin{figure*}[t!]
	\centering
	\includegraphics[width=\textwidth]{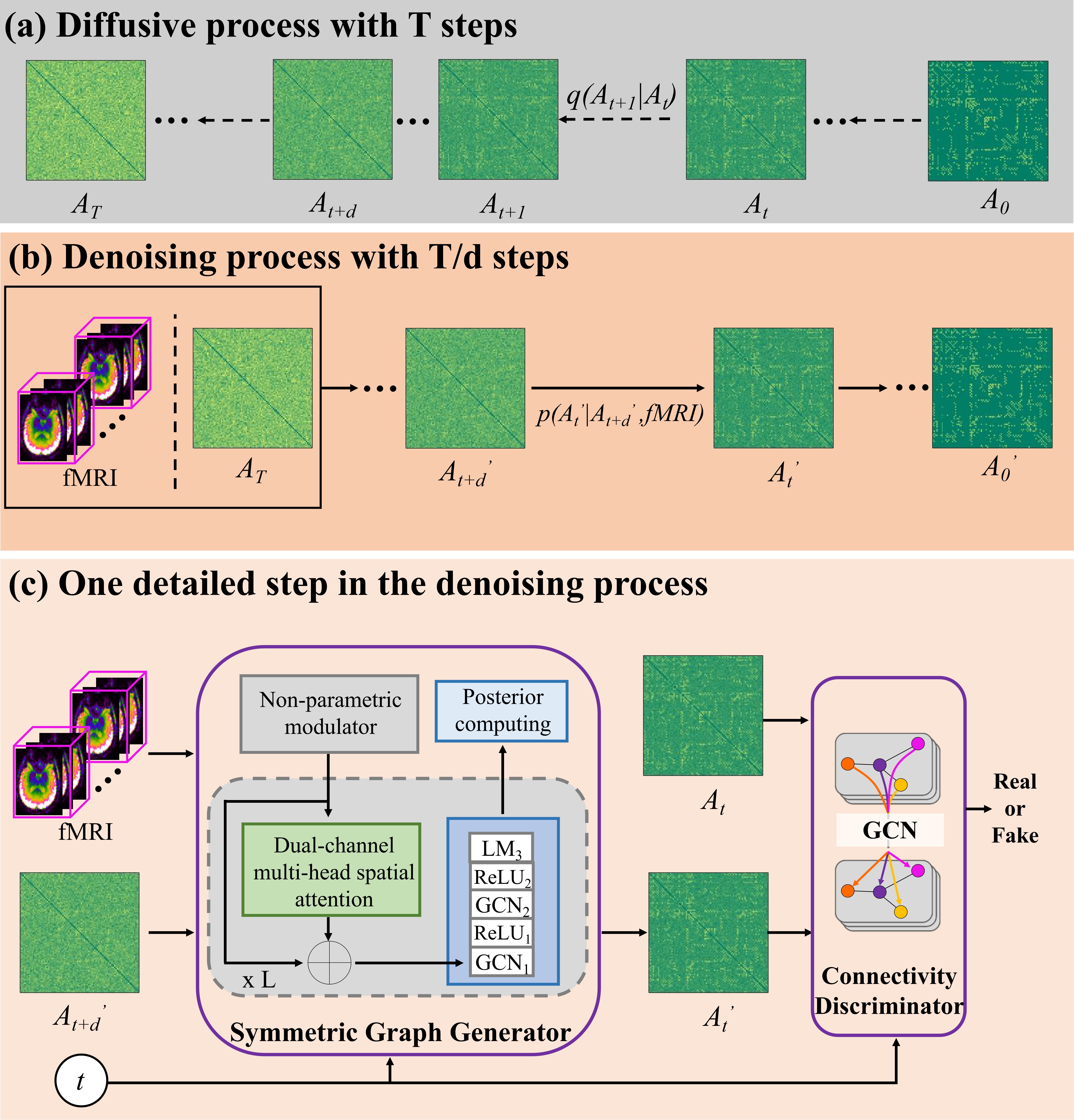}
	\caption{The structure of the proposed DiffGAN-F2S. (a) The empirical SC $A_0$ is diffused to the Gaussian matrix $A_T$ with T steps. (b) The fMRI is the condition and guides the denoising process from the Gaussian matrix $A_T$ to the predicted SC $A_0'$. (c) the details of the generative adversarial network from the $t+d$-th and $t$-th denoising steps.\label{fig2}}
\end{figure*}

\section{Method}
\label{s3}
The architecture of the proposed model is shown in Fig.~\ref{fig2}. It consists of two parts: the diffusive process and the denoising process. The DiffGAN-F2S first models the truth distribution of noisy SC at different steps by persistently injecting Gaussian noise into empirical SC ($\boldsymbol{A}_0$). After T steps, the empirical $\boldsymbol{A}_0$ can be transmitted into the Gaussian matrix $\boldsymbol{A}_T$. Secondly, the brain fMRI is involved in the denoising process as a condition to help predict clean SC $\boldsymbol{A}_0'$. It should be stressed that each denoising step is composed of a generator and a discriminator (Fig.~\ref{fig2}(c)). During the model training, three kinds of loss functions are devised to optimize the generator for precise cross-modal connectivity prediction, including the denoising adversarial loss, the spatially connected consistency loss, and the mean absolute error (MAE) loss.  During the diffusive and denoising processes, the sample $\boldsymbol{A}_i$ is a symmetric matrix with the size $N \times N$, $N$ is the number of brain regions.

\subsection{Symmetric diffusive process}
The diffusive process of Denoising Diffusion Probabilistic Models (DDPM) involves a fascinating approach to generating a Gaussian matrix by gradually transforming empirical SC into Gaussian noise through a sufficient number of T steps. At each step, a weight is applied to the Gaussian noise to establish a Markov chain. These weights are predefined as a variance schedule: $\beta_{1}, \beta_{2}, ..., \beta_{T}$. To simplify the process, the intermediate noisy SC $\boldsymbol{A}_{t+1}$ is computed from the previous noisy matrix $\boldsymbol{A}_{t}$ at step $t$ using the following formulas:

\begin{equation}\label{eq1}
	\boldsymbol{A}_{t+1}=\sqrt{1-\beta_{t}} \boldsymbol{A}_{t}+ \frac{\sqrt{\beta_{t}}}{2} (\boldsymbol{\epsilon} + \boldsymbol{\epsilon}^s), \quad \boldsymbol{\epsilon} \sim \mathcal{N}(\mathbf{0}, \boldsymbol{I})
\end{equation}
where $\boldsymbol{\epsilon}^s$ represents the transpose of $\boldsymbol{\epsilon}$, and $\mathcal{N}$ denotes the Gaussian distribution. The diagonal element of $\boldsymbol{\epsilon}$ is set to 0. The second term to the right of the equal sign is used to generate values in the range of $0\sim1$.$\beta_{t} \in (0,1)$ is defined the same as previous works\cite{ho2020denoising}. The specific parameter settings of the variance schedule are: The initial value of $\beta$ is 0.0001, and the final value of $\beta$ is 0.02. The variance schedule follows a linear interpolation between the initial and final values. This linear scheduling ensures a smooth and gradual injection of noise, avoiding abrupt distribution shifts that could disrupt the diffusion-denoising cycle. The noisy matrix $\boldsymbol{A}_T$ at the last diffusion step is given:
\begin{equation}\label{eq2}
\boldsymbol{A}_{T}=\sqrt{\tilde{\eta}_{t}} \boldsymbol{A}_{0}+\frac {\sqrt{1-\tilde{\eta}_{t}}}{2} ( \boldsymbol{\epsilon}_{t} + \boldsymbol{\epsilon}_{t}^s)
\end{equation}
here, $\eta_{t}=1-\beta_{t}, \tilde{\eta}_{t}=\prod_{j=1}^{t} \eta_{j}$, $\boldsymbol{\epsilon}_{t}$ means the sampled noise matrix from $\mathcal{N}(\mathbf{0}, \boldsymbol{I})$ at t step. Setting a large value for $T$ guarantees that the noisy sample $\boldsymbol{A}_{t}$ closely approximates a Gaussian distribution.

\subsection{Symmetric denoising process}
The aim of our work is to predict individual SC from brain fMRI; the traditional denoising process of DDPM cannot achieve this goal. Therefore, we use the fMRI as a condition to guide the denoising process. Besides, large $T$ needs much more computing time and lowers the SC prediction efficacy. We reduce the denoising steps into $T/d$ steps, where each step is modeled with a conditional generative adversarial network. Here, $d$ represents the skipping steps. Specifically, the symmetric graph generator $\boldsymbol{G}_{\theta}(\boldsymbol{A}_{t+d}',\boldsymbol{fMRI},t)$ accepts the noisy sample $\boldsymbol{A}_{t+d}'$, the conditional $\boldsymbol{fMRI}$, the $t$-th learnable embedding $\boldsymbol{e}_t^g$, and predicts $A_t' \sim p(A_t'|A_{t+d}',fMRI)$. These embeddings are added as biases onto latent feature maps and control the generation difference during denoising steps. Meanwhile, the connectivity discriminator $\boldsymbol{D}_{\phi}(\boldsymbol{A}_{t} or \boldsymbol{A}_{t}',t)$ measures the connectivity distribution difference between real (empirical) $q(\boldsymbol{A}_{t}|\boldsymbol{A}_{0},t)$ and fake (predicted) $p(\boldsymbol{A}_{t}' | \boldsymbol{A}_{t+d}',\boldsymbol{fMRI}))$. The temporal embedding $\boldsymbol{e}_t^d$ also involved the discriminating computation as a bias term. The detailed structure of the generator and discriminator is shown in Fig.~\ref{fig2}(c).

\subsubsection{Symmetric graph generator}
The non-parametric modulator (NPM) is used to transform the raw brain fMRI into preliminary ROI-based time series. As the anatomical atlas (i.e., Anatomical Automatic Labeling, AAL90) divides the whole brain into 90 regions of interest and defines the location and volume for each ROI, we resample each 3D volume of fMRI into the same size of the AAL90 atlas and apply a pixel-level dot product between them. The mean pixel value among the same ROI comprises one point of the ROI-based time series. The whole procedure has no learnable parameters. The output of this modulator is defined as the primary ROI-based time series $\boldsymbol{F}_T=NPM(\boldsymbol{fMRI})$. There are three linear mapping (LM) layers used in our model, we denoted them as $LM_1$, $LM_2$, $LM_3$, respectively.

\begin{figure}[t]
	\centering
	\includegraphics[width=\columnwidth]{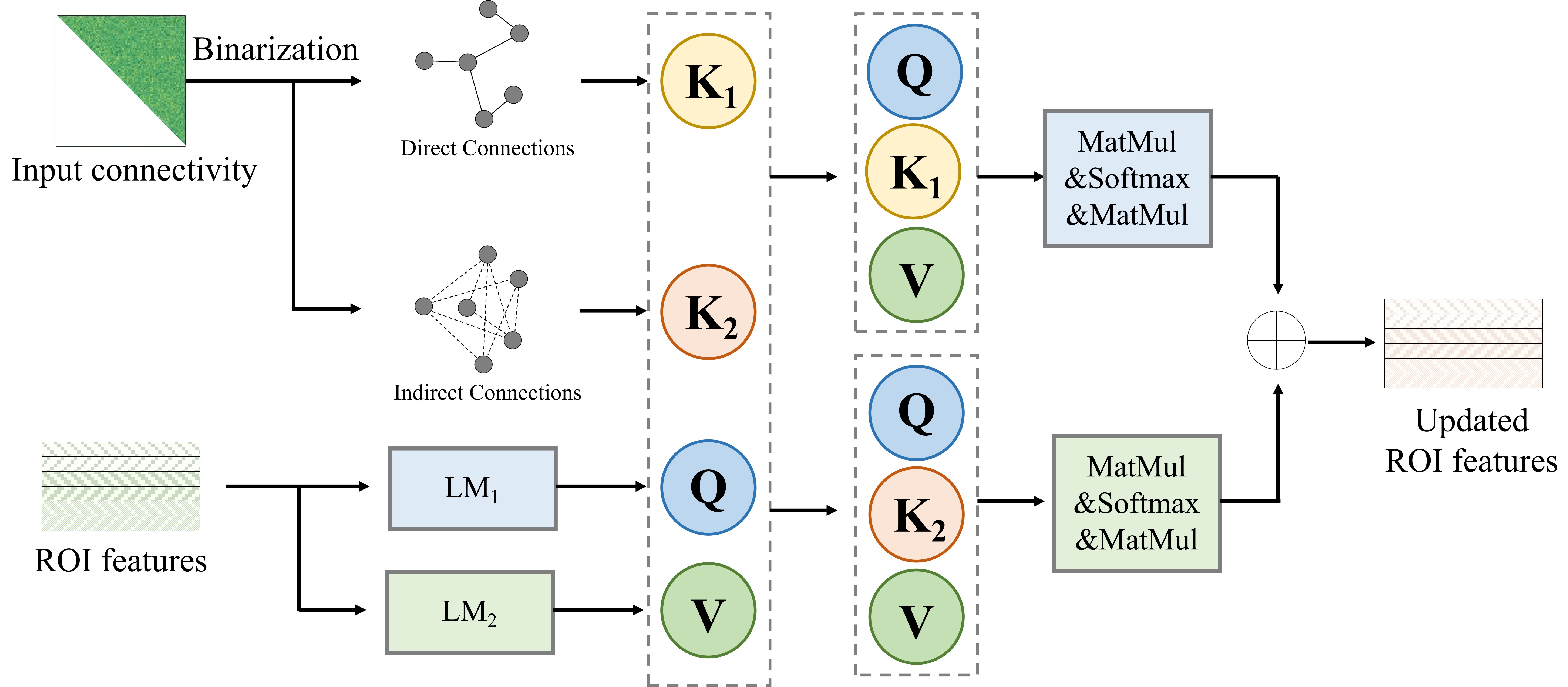}
	\caption{The structure of the dual-channel multi-head spatial attention module. The input is the noisy connectivity and ROI features, and the output is the updated ROI features.\label{fig3}}
\end{figure}

The $\boldsymbol{F}_{t+d}$ and $\boldsymbol{A}_{t+d}'$ pass through $L$ layers consisting of one dual-channel multi-head spatial attention (DMSA) module and one GCN-based module to predict blurred SC ($\boldsymbol{\dot{A}}_0$) at the $t+d$ step. As shown in Fig.~\ref{fig3}, the DMSA captures global relations among directly and indirectly connected brain regions.
The threshold of ROI-to-ROI connection strength is crucial for dividing direct/indirect brain region links, and its selection is primarily based on the references\cite{han2023resting,penalba2023increased}.
The specific implementation involves three steps: first, computing individual FC matrices from ROI-based time series via Pearson correlation (ranging from -1 to 1);
second, retaining only ROI pairs with absolute FC strength greater than 0.3 to exclude weak or spurious correlations;
third, the final binarized matrix $B$ is defined as $B_{ij}=1$ if $FC_{ij} \geq 0.3$, otherwise 0.
For the convenience of description, we specify one layer, denote the input connectivity as $\boldsymbol{A}_{t+d}'$, and denote ROI features as $\boldsymbol{F}_{t+d}$. We compute the Query ($\boldsymbol{Q}$), Key ($\boldsymbol{K}_1$,$\boldsymbol{K}_2$), and Value ($\boldsymbol{V}$) for the ROI features. The $\boldsymbol{Q}$ and $\boldsymbol{V}$ are computed through the linear mapping (LM). The $\boldsymbol{K}_1$ and $\boldsymbol{K}_2$ stand for direct and indirect connected ROI features, which are derived from one-layer GCN. When computing the attention values, the directly connected ROI Keys are combined with Queries to obtain attention maps. We denote the $\cup$ and $\bar{\cup}$ as direct and indirect connected brain region containers. The computation of these modes is expressed by:
\begin{equation}
	\begin{aligned}
		\boldsymbol{F}^{dmsa} &=Att_1(\boldsymbol{Q},\boldsymbol{K_1}) \boldsymbol{V} + Att_2(\boldsymbol{Q},\boldsymbol{K_2}) \boldsymbol{V} + \boldsymbol{F}_{t+d} \\
		&=\operatorname{softmax}\left(\frac{\boldsymbol{Q} \boldsymbol{K}_1^s}{\sqrt{Num(\cup)}}\right) \boldsymbol{V} \\
        &+ \operatorname{softmax}\left(\frac{\boldsymbol{Q} \boldsymbol{K}_2^s}{\sqrt{Num(\bar{\cup})}}\right) \boldsymbol{V} + \boldsymbol{F}_{t+d}
	\end{aligned}
\end{equation}
\begin{equation}
    \begin{aligned}
	\boldsymbol{F}^{graph} &= LM_3(ReLU_2(GCN_2(ReLU_1(GCN_1( \boldsymbol{F}^{dmsa}) \\
                           &+ \boldsymbol{e}_t^g))+ \boldsymbol{e}_t^g)+ \boldsymbol{e}_t^g)
    \end{aligned}
\end{equation}
To be specific, for the $i$-th brain region, the directly connected brain container is denoted as $\cup_i$. The attention value between the $i$-th and the $j$-th brain regions can be computed by:
\begin{equation}
	Att_{1,ij}=\frac{\boldsymbol{V}_{ij} \boldsymbol{K}_{1,ij} / \sqrt{Num(\cup)}}{\sum_{k \in \cup_{i}} \left(\boldsymbol{V}_{ij} \boldsymbol{K}_{1,ij} / \sqrt{Num(\cup)}\right)}
\end{equation}
This formula can also be applied to the calculation of attention value in brain regions that are indirectly connected.

The posterior computing module (PCD) first translates the ROI-based features $\boldsymbol{F}^{graph} $ into symmetrically blurred $\boldsymbol{\dot{A}}_0$, then predicts noisy SC at the $t$-th step by posterior sampling strategy. The computing formula is given:
\begin{equation}
	\boldsymbol{\dot{A}}_0 = \sigma (\boldsymbol{F}^{graph} ( \boldsymbol{F}^{graph})^s ) - \boldsymbol{I}
\end{equation}
\begin{equation}
	\boldsymbol{A}_t' = \sqrt{\tilde{\eta}_{t}} \boldsymbol{\dot{A}}_0 +\frac {\sqrt{1-\tilde{\eta}_{t}}}{2} ( \boldsymbol{\epsilon}_{t} + \boldsymbol{\epsilon}_{t}^s)
\end{equation}
The output $\boldsymbol{A}_t'$ is then used to predict $\boldsymbol{A}_{t-d}'$. $\boldsymbol{E}$ is an identity matrix. $\sigma$ is the sigmoid activation function. After $T/d$ steps, we can obtain the final clean SC $\boldsymbol{A}_0'$.

\subsubsection{Connectivity discriminator}
The discriminator aims to distinguish whether the noisy SC comes from the generator or the empirical method. The network structure is designed with three GCN layers, where the edge is the noisy SC ($\boldsymbol{A}_t$, or $\boldsymbol{A}_t'$) , and the node feature is defined with a one-hot vector.
Consistent with the number of brain regions partitioned by the AAL90 atlas, the dimension of each one-hot vector is 90. This ensures a one-to-one correspondence between each brain region and a unique vector dimension. We strictly follow the anatomical labeling order of the AAL90 atlas for the mapping. The AAL90 atlas partitions the brain into 90 regions of interest (ROIs) with a predefined anatomical order. Using one-hot vectors as node features ensures that each brain region is uniquely identified in the discriminator, avoiding feature ambiguity caused by shared representations.
For each GCN layer, the temporal embedding ($\boldsymbol{e}_t^d$) is inserted as a bias. After three GCN layers, the ROI feature is averaged and mapped into one scalar. The output value of 0 or 1 indicates the fake (predicted) or real (empirical) SC.

\subsection{Loss function}
To enhance the quality of the predicted SC, a hybrid objective function is devised for model optimization. Firstly, the generative adversarial loss constrains the predicted SC and the empirical SC in distribution consistency by computing the discriminative difference between them.
\begin{equation}\label{eq8}
	\begin{array}{r}
		\mathcal{L}_{D}=\frac{d}{T} \sum_{t \geq 1} \mathbb{E}_{q\left(\mathbf{A}_{t} \mid \mathbf{A}_{0}\right)}\left[\left(D_{\phi}\left(\mathbf{A}_{t}, t\right)-1\right)^{2}\right] \\
		+\mathbb{E}_{p_{\theta}\left(\mathbf{A}_{t}' \mid \mathbf{A}_{t+d}'\right)}\left[D_{\phi}\left(\mathbf{A}_{t}', t\right)^{2}\right]
	\end{array}
\end{equation}
\begin{equation}\label{eq9}
	\begin{array}{r}
		\mathcal{L}_{G}= \frac{d}{T} \sum_{t \geq 1} \mathbb{E}_{p_{\theta}\left(\mathbf{A}_{t}' \mid \mathbf{A}_{t+d}'\right)}\left[ (D_{\phi}\left(\mathbf{A}_{t}', t\right) -1 )^{2}\right]
	\end{array}
\end{equation}
Besides, to enforce an additional element-wise constraint on the generator, the MSE loss is added to the adversarial loss by measuring the edge strength difference between the predicted SC and the empirical SC. This loss can make the generator stable when predicting SC. The MSE loss is defined as follows:
\begin{equation}\label{eq10}
	\mathcal{L}_{MSE}= \frac{d}{T} \sum_{t \geq 1} || \boldsymbol{\dot{A}}_{0,t} - \mathbf{A}_{0} ||
\end{equation}

Furthermore, to capture the local and global graphical properties, the spatially connected consistency loss is devised to guide the denoising process to accurately predict intrinsic structural patterns.
\begin{equation}\label{eq11}
	\begin{aligned}
	\mathcal{L}_{SCC}= &\frac{\sum (\boldsymbol{A}_0' - \bar{\boldsymbol{A}_0'})\cdot(\boldsymbol{A}_0 - \bar{\boldsymbol{A}_0})} { \sqrt{\sum (\boldsymbol{A}_0' - \bar{\boldsymbol{A}_0'})^2} \cdot \sqrt{\sum (\boldsymbol{A}_0 - \bar{\boldsymbol{A}_0})^2}  } \\
	&+ \sum_{k} || \sum_{i\neq k \neq j, \boldsymbol{A}_0'} \frac{\kappa_{ij}(k)}{\kappa_{ij}} - \sum_{i\neq k \neq j, \boldsymbol{A}_0} \frac{\kappa_{ij}(k)}{\kappa_{ij}} ||
	\end{aligned}
\end{equation}
Here, the first term measures the overall similarity between predicted and empirical SC, and the second term computes the betweenness centrality (BC) to measure the local topological similarity. $\kappa_{ij}(k)$ is the number of shortest paths from ROI $i$ to ROI $j$ passing through the $k$-th ROI, $\kappa_{ij}$ represents the total number of shortest paths between ROI $i$ and ROI $j$.

The optimization strategy of the proposed DiffGAN-F2S is illustrated using pseudo-code. The Algorithm~\ref{Algorithm1} and Algorithm~\ref{Algorithm2} display the training process and sampling process, respectively.

\begin{algorithm}[h]
	\caption{Training of DiffGAN-F2S}
	\label{Algorithm1}
	\begin{algorithmic}[1]
		\Require
		$\boldsymbol{A}_0$: empirical SC\newline
		\hspace*{1.2em} $\boldsymbol{fMRI}$: conditional imaging\newline
		\hspace*{1.2em} $\boldsymbol{G}_{\theta}$: symmetric graph generator with parameter $\theta$\newline
		\hspace*{1.2em} $\boldsymbol{D}_{\phi}$: connectivity discriminator with parameter $\phi$\newline
		\hspace*{1.2em} $T$: diffusion step number\newline
		\hspace*{1.2em} $\beta_{i},i=1,2,...,T$: predefined variance schedle\newline
		\hspace*{1.2em} $d$: skipping step number
		\Ensure
		$\boldsymbol{G}_{\theta}$ and $\boldsymbol{e}^g$
		\State initialize the $\theta$ and $\phi$, the temporal embedding $\boldsymbol{e}^g$ and $\boldsymbol{e}^d$;
		\Repeat
		\State $t \sim Uniform({d,2d,...,T})$
		\State Sample $\boldsymbol{\epsilon}_{t} \sim \mathcal{N}(\mathbf{0}, \boldsymbol{I})$
		\State Compute the $\boldsymbol{A}_{t}$ and $\boldsymbol{A}_{t+d}$ using Eq.(~\ref{eq1})
		\State Propagate the tuple ($\boldsymbol{A}_{t+d},\boldsymbol{fMRI},\boldsymbol{e}_t^g$) to $\boldsymbol{G}_{\theta}$ to \newline
		\hspace*{1.2em} obtain predicted SC $\boldsymbol{\dot{A}}_{0,t}$ and noisy SC $\boldsymbol{A}_t'$
		\State Propagate the tuple ($\boldsymbol{A}_t, \boldsymbol{e}^d$) or ($\boldsymbol{A}_t', \boldsymbol{e}^d$) to $\boldsymbol{D}_{\phi}$ to \newline
		\hspace*{1.2em} predict true or false
		\State Compute the $\mathcal{L}_{D}$ using Eq.(~\ref{eq8}) and update $\phi$ by back\newline
		\hspace*{1.2em} propagating the gradient $-\nabla_{\phi} \mathcal{L}_{D}$
		\State Compute the combination of Eq.(~\ref{eq9}),Eq.(~\ref{eq10}),Eq.(~\ref{eq11}), \newline
		\hspace*{1.2em} and update $\phi$ by back-propagating the gradient \newline
		\hspace*{1.2em} $-\nabla_{\theta} (\mathcal{L}_{G}+\mathcal{L}_{MSE}+\mathcal{L}_{SCC})$
		\Until converged
	\end{algorithmic}
\end{algorithm}

\begin{algorithm}[h]
	\caption{Inference procedure of DiffGAN-F2S}
	\label{Algorithm2}
	\begin{algorithmic}[1]
		\Require
		$\boldsymbol{fMRI}$: conditional imaging\newline
		\hspace*{1.2em} $\boldsymbol{G}_{\theta}$: symmetric graph generator with parameter $\theta$\newline
		\hspace*{1.2em} $T$: number of diffusion steps\newline
		\hspace*{1.2em} $\boldsymbol{e}^g$: temporal embedding of the generator
		\Ensure
		Clean SC $\boldsymbol{A}_0'$
		\State Sample Gaussian SC $\boldsymbol{A}_{T} = (\boldsymbol{\epsilon} + \boldsymbol{\epsilon}^s)/2$,  $\boldsymbol{\epsilon} \sim \mathcal{N}(\mathbf{0}, \boldsymbol{I})$
		\For {$t=T-d$ to $0$}
		\State Forward-propagate the tuple ($\boldsymbol{A}_{t+d},\boldsymbol{fMRI},\boldsymbol{e}_t^g$) to \newline
		\hspace*{1.2em} $\boldsymbol{G}_{\theta}$ to compute noisy SC $\boldsymbol{A}_{t}'$
		\State $t = t - d$
		\EndFor
		\State Return $\boldsymbol{A}_0'$
	\end{algorithmic}
\end{algorithm}

\begin{figure*}[htbp]
	\centering
	\includegraphics[width=0.9\textwidth]{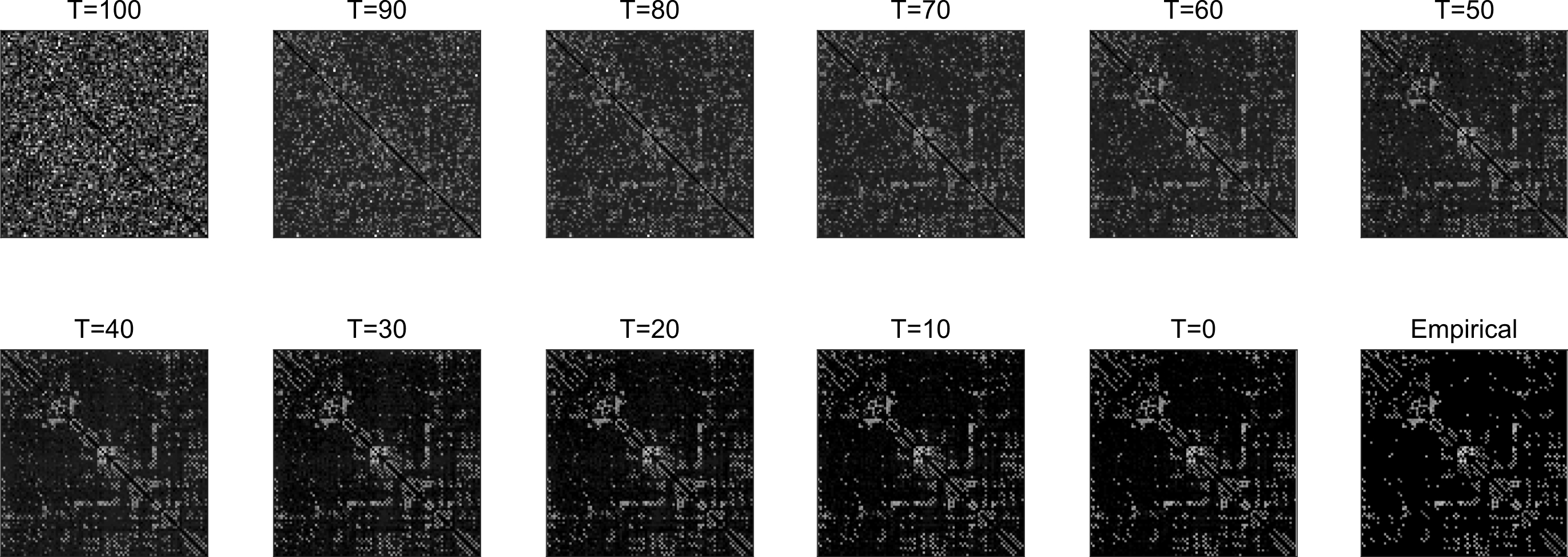}
	\caption{Predicted structural connectivities at different diffusive steps by the proposed model.\label{fig4}}
\end{figure*}

\begin{figure*}[t]
	\centering
	\includegraphics[width=\textwidth]{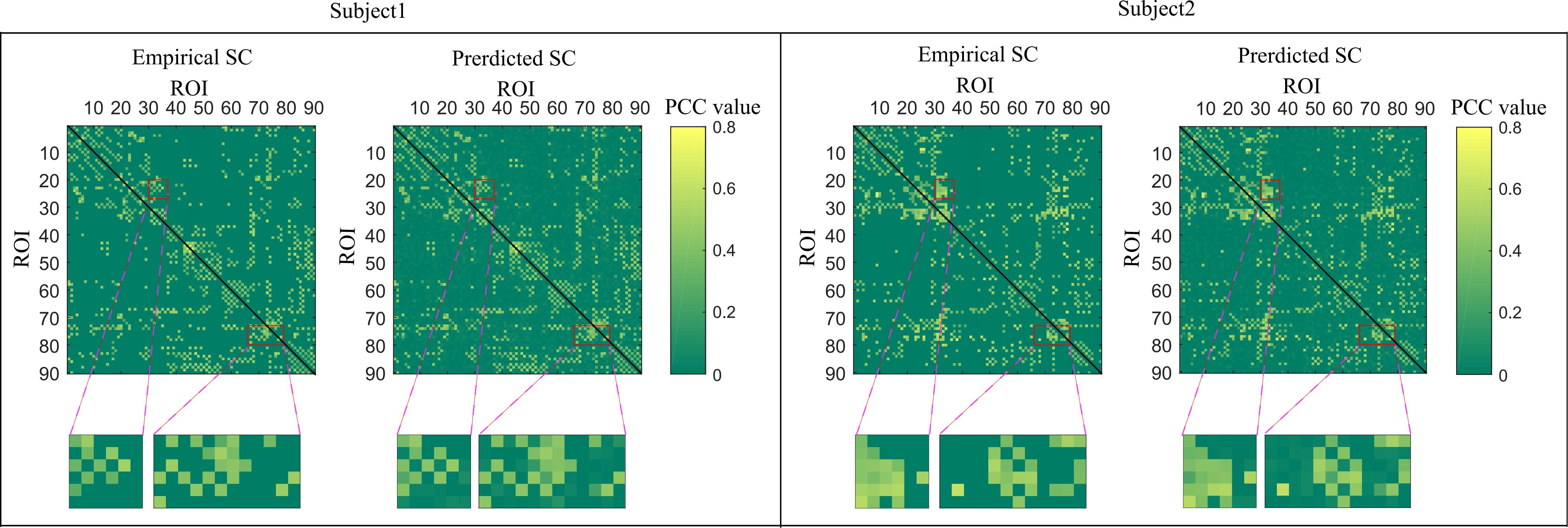}
	\caption{Comparison of the predicted SCs and empirical SCs selected from two subjects.\label{fig5}}
\end{figure*}

\begin{figure}[htbp]
	\centering
	\includegraphics[width=\columnwidth]{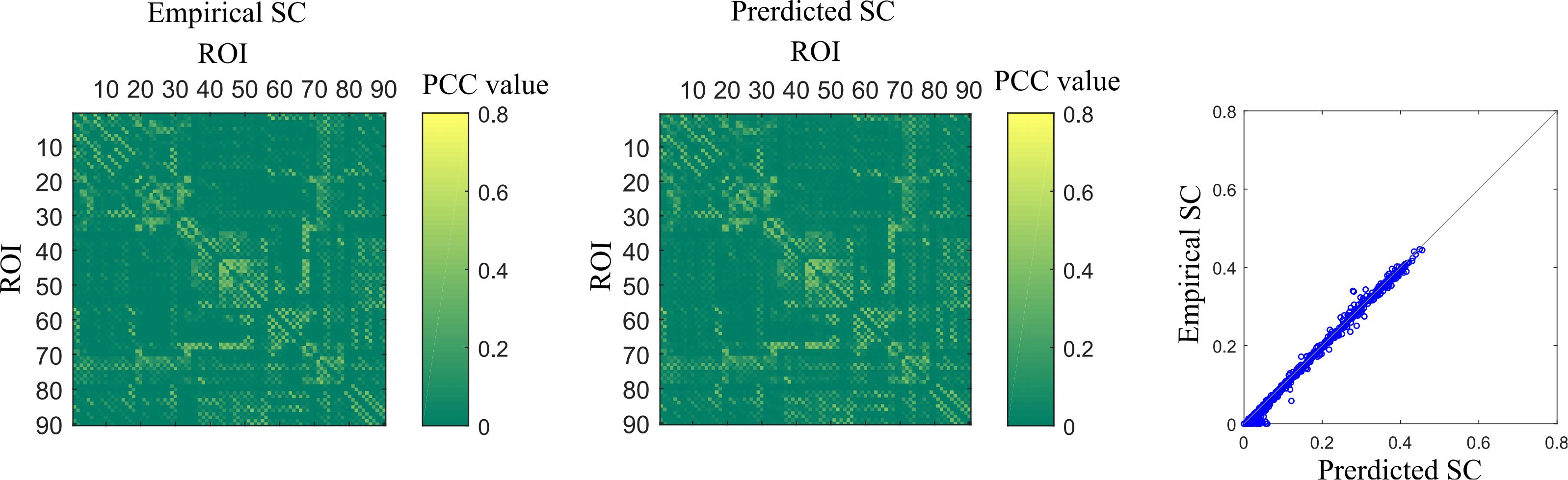}
	\caption{Comparison of group SCs between empirical and predicted SCs. The right column demonstrates the highly correlation between the empirical and predicted SCs. \label{fig6}}
\end{figure}

\section{Experiments}
\label{s4}

\subsection{Datasets}
To test our model's performance, we selected two categories of 240 patients (i.e., normal control and mild cognitive impairment) from the publicly available Alzheimer's Disease Neuroimaging Initiative (ADNI) dataset. Each category has 120 subjects. Mild cognitive impairment (MCI) includes early MCI and late MCI. Every subject was scanned using fMRI and DTI by a 3T magnetic resonance instrument. The DTI is preprocessed by the PANDA\cite{cui2013panda} toolbox. Based on the anatomical automatic labeling (AAL90) atlas \cite{tzourio2002automated}, the detailed preprocessing procedures are described in the work \cite{lei2023multi}. The output of preprocessing DTI is the empirical SC ($\boldsymbol{A}_0$) with the size of $90 \times 90$. For the fMRI preprocessing, we abandon the preprocessing procedures \cite{zuo2021multimodal} and utilize an anatomical atlas file $aal.nii$ to transform the brain fMRI into ROI-based time series without any parameters. The output of fMRI preprocessing is a primary sample $\boldsymbol{F}$ with a size of $90 \times 187$.
The 90 corresponds to the number of ROIs defined by the AAL90 atlas, so each element (i,j) in the $90 \times 187$ matrix represents the mean signal intensity of the $i$-th ROI at the $j$-th time point.
For functional magnetic resonance imaging (fMRI) data, the repetition time (TR) varied from 0.607 s to 3.0 s, and the echo time (TE) was in the range of 30 ms to 32 ms. For diffusion tensor imaging (DTI) data, the TR ranged from 3.4 s to 17.5 s, with the TE spanning 56 ms to 105 ms. Additionally, the number of gradient directions for DTI acquisition was between 6 and 126.

\subsection{Training Settings and Evaluation Metrics}

The aim of our model is to transform the fMRI into structural connectivity step by step. During the model training, we utilize the following parameters: $L=3$, $T=100$. The skipping step number $d$ is set as 10. The code is written using the Pycharm tool and runs on the Windows 11 operating system. The total training epochs are 1000, with a learning rate of $10^{-3}$. The Adam algorithm with default settings is selected to update the weights of the generators and discriminators. The batch size is set to 64. To avoid the bias of unbalanced class distribution, we employ the stratified 5-fold cross-validation to partition the data, which preserves the class distribution (NC:MCI = 1:1) in each fold.

The qualitative results are presented through global and local detail maps. Besides, the predicted results are quantitatively evaluated by eight metrics, including the mean absolute error (MAE), the correlation coefficient (CC), the degree error, the strength error, the clustering error, the betweenness error, the local efficiency error, and the global efficiency error. The former two metrics are the traditional image evaluation methods, and the latter six metrics are the commonly used graph-based measurements. Further, we conduct ROI-based and connectivity-based comparisons to examine our model's generation performance.

\begin{figure*}[htbp]
	\centering
	\includegraphics[width=\textwidth]{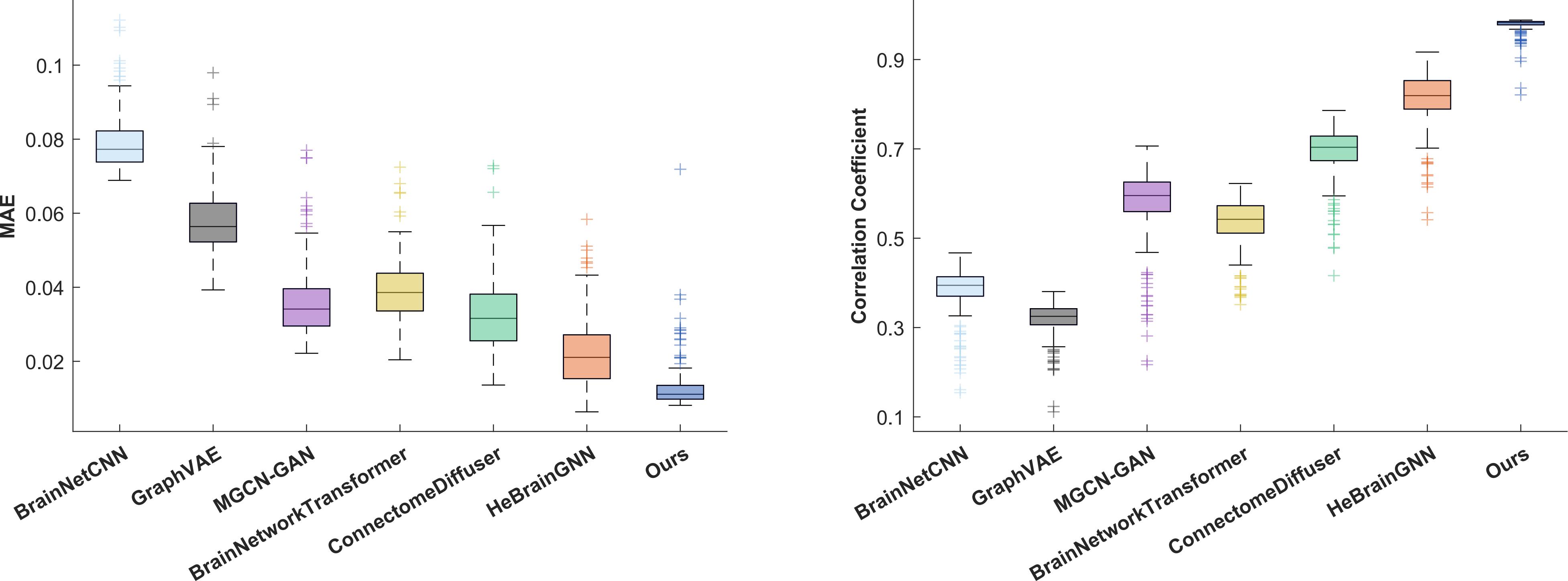}
	\caption{The MAE and correlation coefficient between predicted and empirical SCs using different models.\label{fig7}}
\end{figure*}

\begin{figure*}[htbp]
	\centering
	\includegraphics[width=\textwidth]{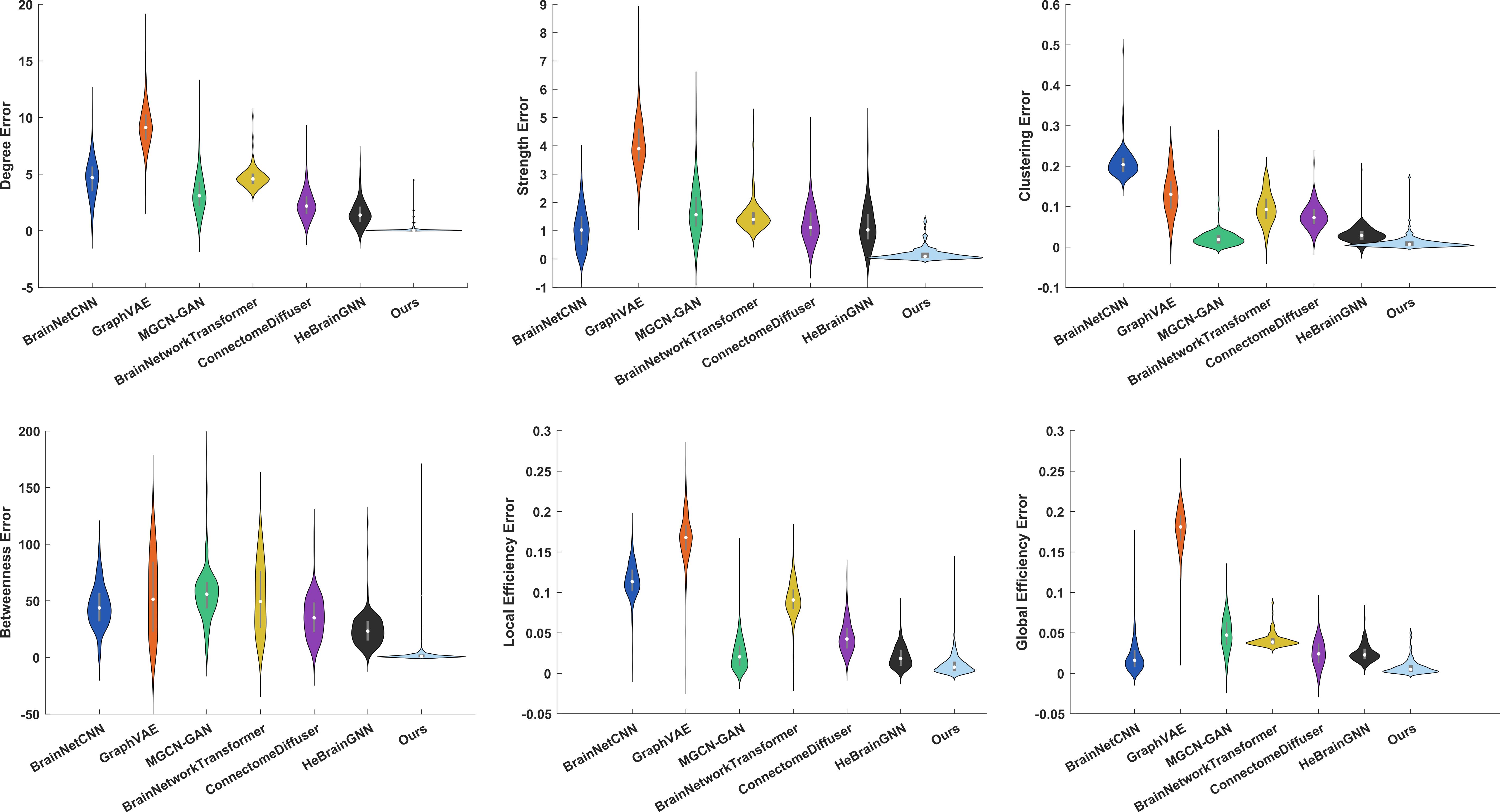}
	\caption{Violin chart comparison of six graph metrics for four different models. The top row is the mean degree error, connectivity strength error and the mean clustering error; the bottom row is the betweenness error, the local efficiency and the global efficiency error. \label{fig8}}
\end{figure*}

\subsection{Denoised and Prediction Results}
In our experiments, given brain fMRI, our model can denoise a symmetrical Gaussian into corresponding structural connectivity. Fig.~\ref{fig4} shows the detailed denoised SCs at 10 representative steps. As the time step decreases to 0, the noise is gradually removed and the empirical connectivity features are preserved. When t = 0, the predicted SC closely resembles the empirical SC. To qualitatively demonstrate the generation performance, two representative subjects are displayed in Fig.~\ref{fig5}. For each subject, both the global and local connectivity patterns are highly preserved. Furthermore, we investigate the group-level connectivity patterns between the empirical and predicted SCs. Specifically, we mean all the predicted SCs and all the empirical SCs, respectively. The subsequent predicted group SC and empirical group SC are reshaped into one-dimensional vectors, which are displayed in a plane. As shown in Fig.~\ref{fig6}, the connectivity element value ranges from 0.0 to 0.8. Each blue circle represents each element in predicted SC and emirical SC. Almost all the blue circles are distributed diagonally. It means there is a high correlation between the empirical and predicted SCs. Also, there are few outliers that deviate from the diagonal line. This means that there are some connections in predicted SCs that are not present in empirical SCs. The possible reason is that the predicted SCs contain little functional-specific information.

\begin{figure}[htbp]
	\centering
	\includegraphics[width=\columnwidth]{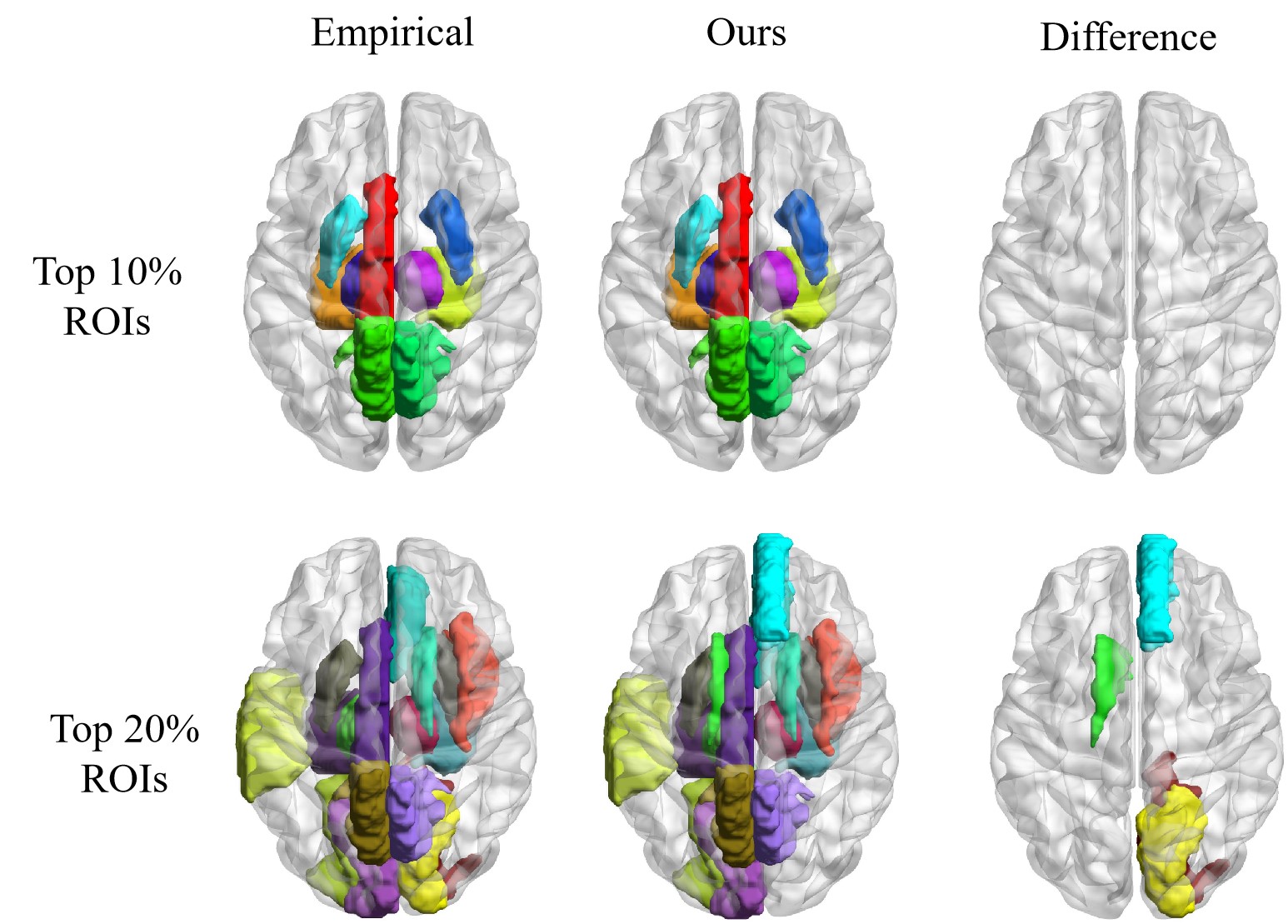}
	\caption{Comparison of the top 10 percent and top 20 percent ROIs between the empirical method and our model; the right column is the ROI distribution difference between them. \label{fig9}}
\end{figure}

To quantitatively evaluate the quality of predicted SCs, we add three other models for comparison. Considering no studies have focused on the tasks of transforming images into graphs, we use functional connectivity (FC) as input in these three models.
(1) The BrainNetCNN model ~\cite{kawahara2017brainnetcnn} with the edge-to-edge convolutional filters;
(2) the graph variational autoencoder (GraphVAE) ~\cite{behrouzi2022graph}, the node features are set as one-hot vectors;
(3) the multi-GCN-based GAN model (MGCN-GAN) ~\cite{zhang2022predicting};
(4) the BrainNetworkTransformer model ~\cite{kan2022brain};
(5) the ConnectomeDiffuser model ~\cite{chen2025connectomediffuser};
(6) the HeBrainGNN model ~\cite{shi2025heterogeneous}.
As shown in Fig.~\ref{fig7}, the boxplots show our model achieves the best performance with the mean MAE value of 0.013 and the mean correlation coefficient of 0.976. The Fig.~\ref{fig8} shows the distribution and density of graph metric errors between predicted SCs and empirical SCs.
The MAE of traditional methods (such as BrainNetCNN, GraphVAE) exceeds 0.05, and the CC is below 0.4. The topological error is significantly high, indicating that these methods have difficulty capturing the fine structure of the brain network. In recent advanced methods (such as ConnectomeDiffuser, HeBrainGNN), although there have been improvements, the MAE is still higher than 0.02, the CC has not exceeded 0.81, and some topological indicators (i.e., degree error, clustering error, and local efficiency error) are still at a relatively high level. The results of our model present a more compact distribution in the box plot (Fig.~\ref{fig7}), and the error distribution in the violin plot (Fig.~\ref{fig8}) is concentrated around 0, indicating that its prediction stability and accuracy are far superior to existing methods. It not only solves the accuracy defects of traditional methods but also compensates for the shortcomings of recent methods in topological consistency.
%Among the six metrics, our model demonstrates the most concentrated distribution around the median value.
This indicates that our model performs the best in the task of generating SC from fMRI among these seven models.

\begin{figure}[htbp]
	\centering
	\includegraphics[width=\columnwidth]{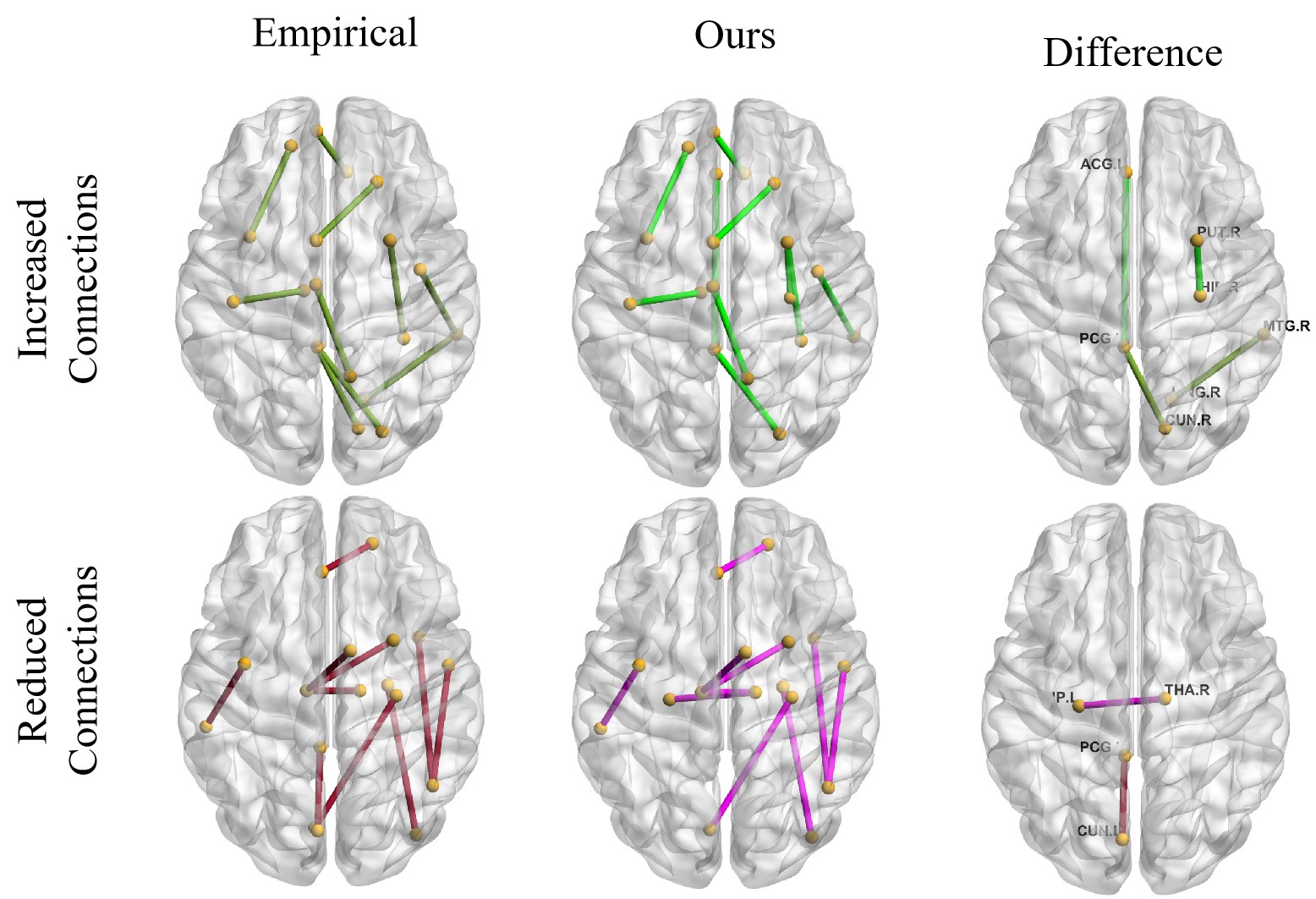}
	\caption{Comparison of the top ten increased and reduced connections between the empirical method and our model; the different connections are displayed in the right column. \label{fig10}}
\end{figure}

\subsection{Analysis of Connectivity}
The connectivity features are important for analyzing brain disease. In our experiment, we investigate the connectivity-based analysis between predicted SCs and empirical SCs. Usually, we first compute the average SC for each group (i.e., NC and MCI) and then subtract the average SC of NC from the average SC of MCI. The different elements are the abnormal connections. Analyzing these abnormal findings can help improve disease diagnosis accuracy and detect potential biomarkers. To compare the important ROIs, we firstly compute the group-averaged SC matrix for the NC group and the MCI group, and then calculate the difference of these two matrices; next, we sum the absolute values of all elements for each row and each column; finally, we compute the mean values of each ROI and sort them in descending order.
This calculation comprehensively captures the overall alteration of connection strength involving each ROI in the pathological process of MCI, and ROIs with larger values are considered more important for distinguishing between the two groups.
The top 10 percent and top 20 percent ROIs are displayed in Fig.~\ref{fig9}. The top 10 percent of ROIs are the same for both empirical and predicted SCs. Out of the top 20 percent ROIs, there are two ROIs that are not consistent. The CAL.R and LING.R calculated from empirical SCs are not presented in the predicted SCs. These two ROIs are presented at the top 30\% ROIs calculated from the predicted SCs.
To compare the top 10 abnormal connections between NC and MCI, we select the top 10 positive values and the last 10 negative values in the subtracted SC. These connections are displayed in Fig.~\ref{fig10}. Both the empirical method and ours show most of the common connections except one or two connections, which means we can utilize our model to generate SCs and obtain the same analysis results.

\begin{table*}[t!]
	\centering
	\renewcommand\arraystretch{1.4}
	\setlength{\abovecaptionskip}{0pt}%
	\setlength{\belowcaptionskip}{10pt}%
	\caption{The impact of loss function and core modules on the SC prediction performance.\label{tab1}}
	\resizebox{\textwidth}{!}{\begin{tabular}{ccccccccc}
\hline
\multicolumn{1}{l}{}                                                 & MAE                                                          & CC                                                           & \begin{tabular}[c]{@{}c@{}}Degree\\ Error\end{tabular}       & \begin{tabular}[c]{@{}c@{}}Strength\\ Error\end{tabular}     & \begin{tabular}[c]{@{}c@{}}Clustering\\ Error\end{tabular}   & \begin{tabular}[c]{@{}c@{}}Betweenness\\ Error\end{tabular}  & \begin{tabular}[c]{@{}c@{}}Local Efficiency\\ Error\end{tabular} & \begin{tabular}[c]{@{}c@{}}Global efficiency\\ Error\end{tabular} \\ \hline
\begin{tabular}[c]{@{}c@{}}DiffGAN-F2S \\ w/o symmetric\end{tabular} & \begin{tabular}[c]{@{}c@{}}0.014\\ ($\pm$0.008)\end{tabular} & \begin{tabular}[c]{@{}c@{}}0.968\\ ($\pm$0.023)\end{tabular} & \begin{tabular}[c]{@{}c@{}}0.418\\ ($\pm$1.622)\end{tabular} & \begin{tabular}[c]{@{}c@{}}0.166\\ ($\pm$0.312)\end{tabular} & \begin{tabular}[c]{@{}c@{}}0.019\\ ($\pm$0.025)\end{tabular} & \begin{tabular}[c]{@{}c@{}}0.014\\ ($\pm$0.265)\end{tabular} & \begin{tabular}[c]{@{}c@{}}0.018\\ ($\pm$0.021)\end{tabular}     & \begin{tabular}[c]{@{}c@{}}0.010\\ ($\pm$0.009)\end{tabular}      \\ \hline
\begin{tabular}[c]{@{}c@{}}DiffGAN-F2S \\ w/o DMSA\end{tabular}      & \begin{tabular}[c]{@{}c@{}}0.031\\ ($\pm$0.012)\end{tabular} & \begin{tabular}[c]{@{}c@{}}0.927\\ ($\pm$0.039)\end{tabular} & \begin{tabular}[c]{@{}c@{}}2.631\\ ($\pm$3.967)\end{tabular} & \begin{tabular}[c]{@{}c@{}}0.371\\ ($\pm$0.829)\end{tabular} & \begin{tabular}[c]{@{}c@{}}0.114\\ ($\pm$0.037)\end{tabular} & \begin{tabular}[c]{@{}c@{}}0.293\\ ($\pm$0.186)\end{tabular} & \begin{tabular}[c]{@{}c@{}}0.079\\ ($\pm$0.023)\end{tabular}     & \begin{tabular}[c]{@{}c@{}}0.026\\ ($\pm$0.015)\end{tabular}      \\ \hline
\begin{tabular}[c]{@{}c@{}}DiffGAN-F2S \\ with $RALW$\end{tabular} & \begin{tabular}[c]{@{}c@{}}0.023\\ ($\pm$0.010)\end{tabular} & \begin{tabular}[c]{@{}c@{}}0.952\\ ($\pm$0.031)\end{tabular} & \begin{tabular}[c]{@{}c@{}}1.409\\ ($\pm$3.918)\end{tabular} & \begin{tabular}[c]{@{}c@{}}0.294\\ ($\pm$0.682)\end{tabular} & \begin{tabular}[c]{@{}c@{}}0.052\\ ($\pm$0.028)\end{tabular} & \begin{tabular}[c]{@{}c@{}}0.187\\ ($\pm$0.227)\end{tabular} & \begin{tabular}[c]{@{}c@{}}0.039\\ ($\pm$0.017)\end{tabular}     & \begin{tabular}[c]{@{}c@{}}0.018\\ ($\pm$0.012)\end{tabular}      \\ \hline
\begin{tabular}[c]{@{}c@{}}DiffGAN-F2S \\ w/o $L_{SCC}$\end{tabular} & \begin{tabular}[c]{@{}c@{}}0.028\\ ($\pm$0.011)\end{tabular} & \begin{tabular}[c]{@{}c@{}}0.931\\ ($\pm$0.038)\end{tabular} & \begin{tabular}[c]{@{}c@{}}1.985\\ ($\pm$4.187)\end{tabular} & \begin{tabular}[c]{@{}c@{}}0.351\\ ($\pm$0.837)\end{tabular} & \begin{tabular}[c]{@{}c@{}}0.068\\ ($\pm$0.033)\end{tabular} & \begin{tabular}[c]{@{}c@{}}0.286\\ ($\pm$0.210)\end{tabular} & \begin{tabular}[c]{@{}c@{}}0.058\\ ($\pm$0.020)\end{tabular}     & \begin{tabular}[c]{@{}c@{}}0.021\\ ($\pm$0.013)\end{tabular}      \\ \hline
DiffGAN-F2S                                                          & \begin{tabular}[c]{@{}c@{}}0.013\\ ($\pm$0.007)\end{tabular} & \begin{tabular}[c]{@{}c@{}}0.976\\ ($\pm$0.021)\end{tabular} & \begin{tabular}[c]{@{}c@{}}0.416\\ ($\pm$1.623)\end{tabular} & \begin{tabular}[c]{@{}c@{}}0.164\\ ($\pm$0.311)\end{tabular} & \begin{tabular}[c]{@{}c@{}}0.016\\ ($\pm$0.023)\end{tabular} & \begin{tabular}[c]{@{}c@{}}0.010\\ ($\pm$0.261)\end{tabular} & \begin{tabular}[c]{@{}c@{}}0.015\\ ($\pm$0.018)\end{tabular}     & \begin{tabular}[c]{@{}c@{}}0.008\\ ($\pm$0.009)\end{tabular}      \\ \hline
\end{tabular}}
\end{table*}

\subsection{Ablation Study}
The proposed DiffGAN-F2S model aims to denoise the fMRI into structural connectivity. To analyze the influence of the loss functions on generation performance, we remove the adversiral loss and spatially connected consistency loss, respectively. The two variants are (1) DiffGAN-F2S with $RALW$ means DiffGAN-F2S with reduced adversarial loss weight(RALW), we keep the discriminator structure unchanged, and reduce the weight of the adversarial loss in the total loss from 1.0 to 0.2 (other loss weights remain unchanged); and (2) DiffGAN-F2S w/o $L_{SCC}$. The eight metrics introduced are calculated to estimate the generation performance. As shown in Table~\ref{tab1}, the core function of the adversarial loss is to constrain the distribution of the predicted SC to be consistent with the empirical SC. After reducing its weight, the MAE increases by 0.01, and the topological index error significantly increases. This verifies the necessity of the adversarial loss in improving the consistency of the distribution of the generated results.
Also, the removal of $L_{SCC}$ fails to capture the global topological characteristics, thus producing a greater MAE of 0.015 than that of DiffGAN-F2S.

Furthermore, we investigate the effect of three core factors on the model's prediction performance, including (1) effect of symmetric vs. asymmetric design, (2) impact of DMSA vs. standard attention, (3) necessity of the d-step skipping mechanism.
As shown in Table~\ref{tab1}, the experimental results clearly demonstrate the core role of the symmetrical constraint: After removing the symmetrical design, the MAE significantly increased from 0.013 to 0.014, and the correlation coefficient (CC) decreased from 0.976 to 0.968; The deterioration of topological indicators was more obvious.
When replacing the DMSA with standard attention, the MAE of the standard attention variant is 0.018 higher than that of the original model, the CC decreases by 0.049, and the error of topological indicators increases significantly.
We test the model performance under different skipping step sizes (d = 5, 10, 20). The key results are shown in Fig.11. When d=5, our model can achieve almost the same results with that of d=10, but it requires 20 complete denoising steps and the single-sample prediction time is long as 5.39 seconds; When d = 10, the model maintains high accuracy (MAE = 0.013, CC = 0.976) while compressing the prediction time to 2.73 seconds, achieving a balance between accuracy and efficiency; When d = 20, the efficiency is further improved, but the MAE increases by 0.003 and the CC decreases by 0.022. The reason is that the excessive skipping step size leads to information loss during the denoising process and makes it difficult to fully fit the data distribution.

\begin{figure}[htbp]
	\centering
	\includegraphics[width=0.8\columnwidth]{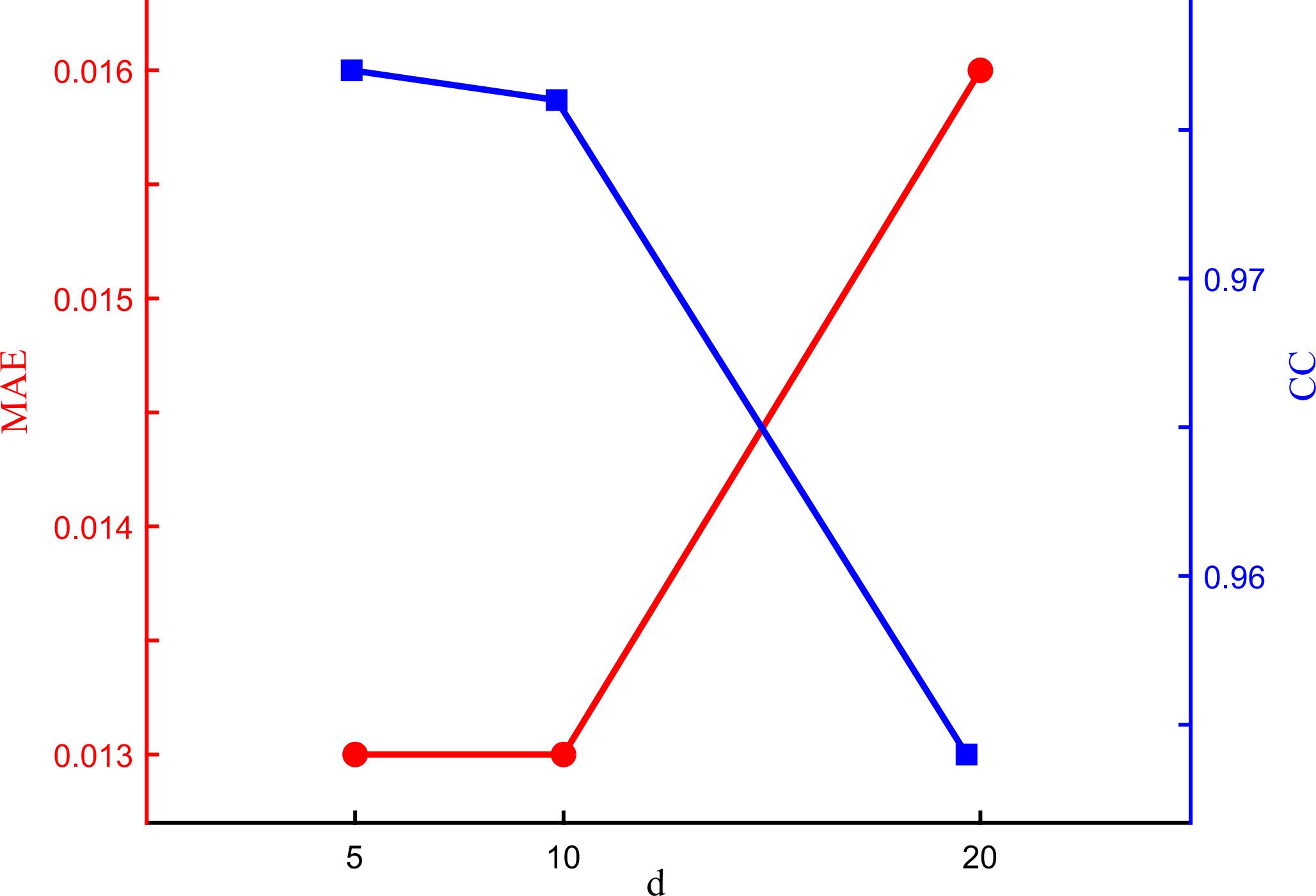}
	\caption{Impact of different skipping step $d$ on the SC prediction performance. \label{fig11}}
\end{figure}

\section{Discussion}
\label{s5}
\begin{figure}[htbp]
	\centering
	\includegraphics[width=0.8\columnwidth]{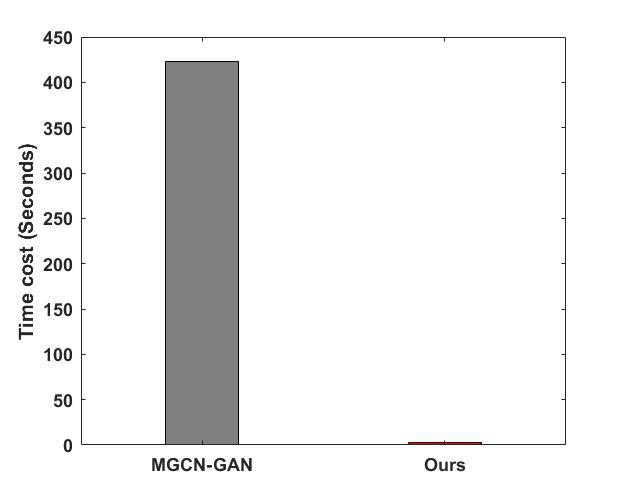}
	\caption{Comparison of SC predicted efficiency between the MGCN-GAN and Ours. \label{fig12}}
\end{figure}

The main difference between our model and previous studies is that we transform the original brain fMRI into structural connectivity in one stage. The proposed model can be easily implemented step by step. The non-parameter transforming step can make use of the previously defined AAL90 atlas to partition each volume into 90 parts for all time points. The subsequent 90 parts can be used as ROI features in the denoising process. At each step of the denoising process, the denoised SC is sent into the discriminator to learn connectivity-consistent patterns with the diffusive SC.
Combining the GAN and DDPM can benefit two advantages: the ability to generate high-quality and diverse SCs and the fast denoising computation. Table~\ref{tab1} shows that the removal of generators can lead to poor performance in fMRI-to-SC prediction. The proposed model generates good SCs that are very similar to empirical SCs. Both subject-specific and group-based qualitative comparisons demonstrate that the predicted SCs by the proposed model can totally capture the empirical connectivity features. Besides, compared with other competing FC-to-SC models, the proposed model can achieve the best results in terms of MAE, correlation coefficient, degree error, strength error, clustering error, betweenness error, local efficiency error, and global efficiency error. In terms of prediction accuracy, our method achieves the best MAE of 0.013, outperforming traditional methods by avoiding FC preprocessing-induced error propagation. For efficiency (as shown in Fig.~\ref{fig12}), our end-to-end pipeline cuts single-SC prediction time from $\sim$420s to $\sim$3s, enabling clinical real-time applications for cross-modal brain network mapping.

\begin{figure}[htbp]
	\centering
	\includegraphics[width=\columnwidth]{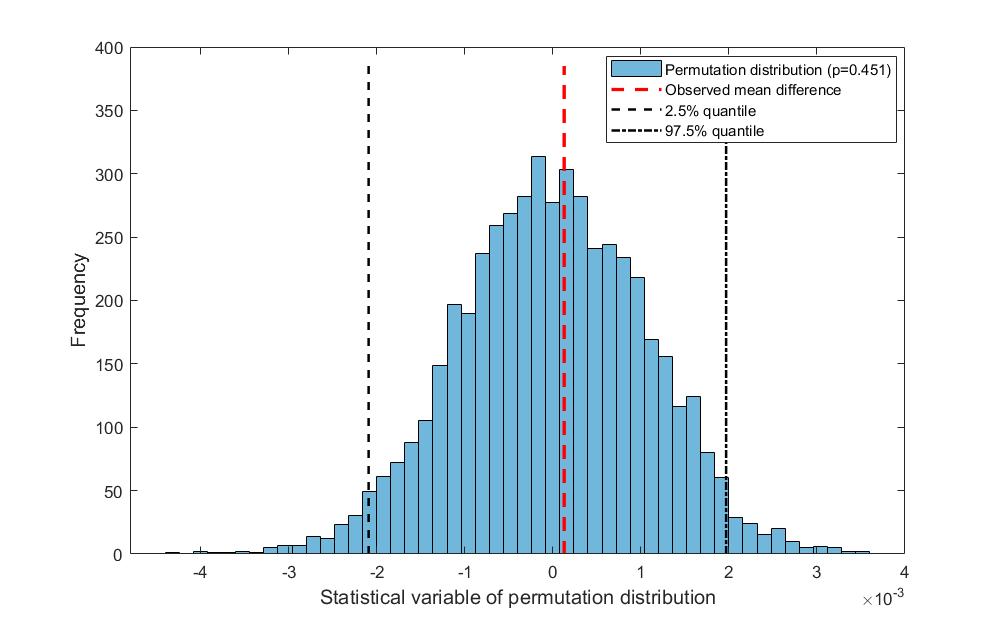}
	\caption{Permutation testing between empirical SC and predicted SC using the proposed model. \label{fig13}}
\end{figure}

To statistically validate our results' effectiveness, we conduct permutation testing between empirical and predicted SCs. Fig.~\ref{fig13} shows a p-value of 0.45, indicating no significant difference between predicted and empirical SC. Our results can capture the real brain structure.
The identified top 10\% ROIs are completely consistent between the empirical method and our method, and 80\% of the top 20\% ROIs overlap validate that our model can reliably capture disease-relevant brain regions. The overlapping regions include the left dorsal cingulate gyrus (DCG.L), bilateral hippocampus (HIP.L/R), bilateral precuneus (PCUN.L/R), bilateral putamen (PUT.L/R), and bilateral thalamus (THA.L/R). These regions are related to MCI disease. For example, the hippocampus (HIP.L/R) and precuneus (PCUN.L/R) are critical for episodic memory and spatial cognition, which can affect the progression from MCI to Alzheimer's disease (AD)\cite{hurtz2019automated,xiong2023disrupted}.
The two non-overlapping ROIs identified in the empirical SC but not in the predicted SC's top 20\% (CAL.R, LING.R) have weaker associations with early MCI pathology.
The absence of CAL.R and LING.R in the predicted top-20\% ROIs is not a false negative, but rather a reflection of the model's ability to prioritize disease-relevant, robust features while filtering less consistent, or non-specific MCI-related changes. The CAL.R and LING.R are visual cortex regions, and previous studies report that visual system abnormalities are relatively late manifestations in MCI\cite{kusne2017visual,marquie2019visual}.
Furthermore, the comparison of the top ten increased and reduced connections focuses on the group-level connectivity difference between MCI and NC. The majority of top increased/reduced connections are shared between empirical and predicted SCs, confirming our model's ability to capture clinically meaningful structural changes. The reduced connections reflect weakened neural activity in MCI, and the increased connections reflect compensatory neural activity in MCI.

There are still two limitations in the current study. One is that we rely solely on the AAL90 template for brain parcellation, whose coarse spatial resolution merges functionally distinct sub-regions. Different templates vary in ROI definition, which may lead to inconsistent prediction results. We will incorporate high-resolution atlas to validate our model on SC prediction in future work. Another one is that we ignore the directional connections in cross-modal prediction. The directional information between brain regions can make the predicted results more interpretable, and its absence reduces the biological interpretability of the predicted SCs. In the future, we will explore asymmetric diffusion learning with directional attention mechanisms to model directional structural connections and apply this improved model in other larger datasets.

\section{Conclusion}
\label{s6}
This paper proposes a DiffGAN-F2S model to predict SC from brain fMRI in one stage. The DiffGAN-F2S combines diffusion models and generative adversarial networks to generate high-fidelity SCs through a few denoising steps. Stacking the dual-channel multi-head spatial attention and the GCN-based modules in the symmetric graph generator can capture global relations among direct and indirect connected brain regions. To further preserve global-local topological information, a spatially connected consistency loss is devised to constrain the generator to accurately predict empirical SCs. Experimental results on the public ADNI dataset show the superior generation performance of the proposed model compared to other competing models.
The proposed model can also be used to identify disease-related connections and important brain regions, which provides an effective way for fMRI-to-SC prediction for multimodal brain network fusion.

\section*{Acknowledgement}
This work was supported partly by the National Natural Science Foundation of China (62406107), partly by the Natural Science Foundation of Hubei Province (JCZRLH202600147), partly by the Education Department Scientific Research Program Project of Hubei Province of China (Grant Number Q20232206), partly by the Hubei University of Economics Youth Research Fund Project (XJZD202505).
Data collection in preparation of this work were funded from the Alzheimer's Disease Neuroimaging Initiative (ADNI) database(http://adni.loni.usc.edu/).

\section*{Conflict of Interest Statement}
The authors declare that the research was conducted in the absence of any commercial or financial relationships that could be construed as a potential conflict of interest.

\bibliographystyle{IEEEtran}
\bibliography{ref}

\end{document}